\documentclass{article}
\usepackage{ijcai18}

\usepackage{times}
\usepackage{xcolor}
\usepackage{soul}
\usepackage[utf8]{inputenc}
\usepackage[small]{caption}

\usepackage{CJKutf8}
\usepackage{graphicx}
\usepackage{amsmath}
\usepackage{subfigure}
\usepackage{multirow}
\usepackage[font={small}]{caption}
\title{Generating Thematic Chinese Poetry using \\Conditional Variational Autoencoders with Hybrid Decoders}
\author{Xiaopeng Yang, Xiaowen Lin, Shunda Suo, and Ming Li\\
David R. Cheriton School of Computer Science,\\
Faculty of Mathematics,  University of Waterloo\\
Waterloo, ON, Canada N2L 3G1\\
\{x335yang, x65lin, sdsuo, mli\}@uwaterloo.ca
}

\date{}

\begin{document}
\maketitle
\begin{abstract}
Computer poetry generation is our first step towards computer writing.
Writing must have a theme. The current approaches of using
sequence-to-sequence models with attention often produce non-thematic poems.
We present a novel conditional variational autoencoder 
with a hybrid decoder adding the deconvolutional neural networks to the general recurrent neural networks to fully learn topic information via latent variables.
This approach significantly improves the relevance of
the generated poems by representing each line of the poem not only in a
context-sensitive manner but also in a holistic way that is highly related to
the given keyword and the learned topic.
A proposed augmented word2vec model further improves the rhythm and symmetry.
Tests show that the generated poems by our approach are mostly satisfying with regulated rules and consistent themes, and 73.42\% of them receive an \textit{Overall} score no less than 3 (the highest score is 5).

\end{abstract}

\section{Introduction}
Poetry is a beauty of simplicity. Its abstractness, concise formats, and rules provide
regularities as the first target of language generation. 
Such regularity is especially amplified in the classical Chinese poetry, for example, 
the quatrains where each poem (1) consists of four lines, each with five or seven characters,
(2) the last character in the second and fourth line follow the same rhythm, and (3) tonal pattern
requests characters in particular positions hold particular tones in terms of Ping (level tone)
and Ze (downward tone) \cite{wang2002}.
An example of a quatrain 
written by Bo Wang, a famous poet in the Tang Dynasty, 
is shown in Table \ref{poem:tab1}.
As illustrated in Table \ref{poem:tab1}, a good quatrain should 
follow all the three pattern regularities mentioned above. 

Besides the rules, a poem is an expression of a certain theme or human emotion.
It has to hold consistent semantic meanings and emotional expressions. 
It is not trivial to create a quatrain by certain 
rules of rhythm and tone, and express a consistent theme 
or some consistent feelings even by people. 
Automatically generating poetry that contains what we want it to express
is a primary task of language generation.

\begin{CJK}{UTF8}{gbsn}
\begin{table}
\centering
\begin{tabular}{|c|} \hline
长江悲已滞，(*PPZZ)\\
Long stay by the Yangtze River, \\
万里念将归。(*ZZPP)\\
Thousands of Miles away from home, \\
况属高秋晚，(*ZPPZ)\\
Yellow Leaves in late autumn wind, \\
山中黄叶飞。(PPZZP)\\
Fall and float in hills, make me sad.\\
\hline
\end{tabular}
\caption{
An example of five-character quatrain.
The tonal pattern is shown at the end of each line, where `P' indicates a level tone, `Z' indicates a downward tone, and `*' indicates the tone can be either. 
}
\vspace{-3mm}
\label{poem:tab1}
\end{table}
\end{CJK}


Major progress has been made in poetry generation \cite{he2012generating,bahdanau2014neural,wang2016chinese,zhang2017flexible}.
Even though the existing approaches have shown their great power 
in poetry automatic generation, they still suffer from a major problem:
lack of consistent theme representation and unique emotional expression.
Taking poem shown in Table \ref{poem:tab1} for instance, the consistent theme of 
this poem is nostalgia.
Apparently, every single line of this poem is related closely to the theme and emotion. 
Recent work \cite{wangzhe2016chinese,hopkins2017automatically} have tried to 
generate poems with the smooth and consistent theme by using topic planning scheme or similar word extensions. It is still hard for these methods to represent topics 
and use them to further improve the quality of generated poems. 

In this paper, we try to solve the difficulty in learning the themes of poetry,
meanwhile leveraging them to boost the generation of corresponding 
poems. As Variational AutoEncoders (VAE) \cite{kingma2013auto} 
have been proved effective in topic representation using learned latent variables for 
text generation \cite{bowman2015generating,serban2017hierarchical}, 
we regard VAE as a possible solution. Moreover, since most 
written poems are composed under certain ``intent,''
we seek for Conditional Variational AutoEncoders (CVAE),
a recent modification of VAE, to generate diverse images/texts conditioned on 
certain attributes \cite{yan2016attribute2image,sohn2015learning,zhao2017learning}. In our work, 
we hypothesize a part of the ``intent'' can be represented in the form of keywords as the conditions for VAE, and the other part can be expressed by the latent variables learned from CVAE.
The general CVAE where both the encoder and decoder are RNNs usually faces the \textit{vanishing latent variable problem} \cite{bowman2015generating} when applied directly to natural language generation.
Thus, we present a novel CVAE with a hybrid decoder (CVAE-HD), which contains both deConvolutional Neural Networks (deCNN) and Recurrent Neural Networks (RNN), to fully learn information from the learned latent variables. 
In addition, we propose to add vertical slices of poems as additional sentences in training data for the word2vec model in order to further improve the rhythm and symmetry delivered in poems, and name this as an Augmented Word2Vec model (AW2V).  
We also propose a straightforward and easily applied automatically evaluation metric Rhythm Score Evaluation (RSE) to measure the poetry rule-consistency. 
Specifically, the contributions of this paper can be summarized as follows:
\begin{itemize}
\item We propose to use conditional variational autoencoders to learn the theme information from poetry lines.
To the best of our knowledge, this represents the first attempt at using CVAE for poetry generation.

\item We present a novel conditional variational autoencoder with a hybrid decoder combining deCNN with the general RNN, which demonstrates the capability of learning topic information from poems and also addressing the vanishing latent variable problem. 
\item We introduce an augmented word2vec model to improve the rhythm and symmetry delivered in poems. Experiments show that AW2V is not only able to boost the rule-consistency of generated poems, but also can be used to search characters representing similar semantic meanings in Chinese poems.  
\item We build a Chinese poetry generation system which can take users' writing intent into the generation process. 
The experimental results show that our system can generate good quatrains which satisfy the rules and have a consistent topic. 
\end{itemize}

\section{Related Work}\label{relatedwork}
Poetry generation is our first step toward experimenting language generation.
According to the methodology used in different approaches, 
we categorize those methods into three major directions, i.e., 
approaches using rules/templates, 
 approaches using Statistical Machine Translation (SMT) models 
and approaches using neural networks.

The first kind of approach is based on rules and/or templates, such as 
phrase search \cite{wu2009new}
and genetic search~\cite{zhou2010genetic}.

The second kind of approach involves various statistical machine translation methods. 
Rather than designing algorithms to identify useful rules,
 the approaches using SMT models, 
whose parameters are derived from the analysis of bilingual text corpora, 
regard the previous line of each poem as the source language in the 
Machine Translation (MT) task and the posterior line as 
the target language sentence \cite{jiang2008generating,he2012generating}.

Due to the fact that all the approaches mentioned above are based on the 
superficial meanings of words or characters, they suffer from the lack of 
deep understanding of the poems' semantic meaning. 
To address this issue, many approaches using neural networks have been proposed 
and attracted much attention in recent years. 
For example, \cite{zhang2014chinese} proposed an approach using 
Recurrent Neural Networks (RNN) that generate each new poem line 
character-by-character (see also \cite{hopkins2017automatically}), with all the 
lines generated previously as a contextual input. 
Experimental results show that quatrains of reasonable quality 
can be generated using this approach. Following this RNN-based approach, 
\cite{wang2016chinese} proposed a character-based RNN treating a poem as 
an entire character sequence, which can be easily extended to various genres 
such as Song Iambics. This approach has the advantage of the flexibility 
and easy implementation, but the long-sequence generation process causes the 
instability of poetry theme. 
To avoid this situation, \cite{wang2016chinese} further brought forward 
the attention mechanism~\cite{bahdanau2014neural} into the RNN-based framework, 
and encoded human intention to guide the poetry generation.
\cite{yan2016poet} proposed an RNN-based poetry generation model with an iterative polishing scheme. Specifically, they encoded users' writing intent first and then decoded it using a hierarchical recurrent neural network.
Recently, \cite{zhang2017flexible} proposed a memory-augmented neural model 
trying to imitate poetry writing process. 
This approach uses the augmented memory to refine poems generated via the neural model, which can balance 
the requirements of linguistic accordance and aesthetic innovation to some extent.
Parallel efforts have been made in generating English poems. 
For instance, \cite{hopkins2017automatically} considered 
adding a list of similar words to a key theme.

We follow the third type of approach to automatically generate Chinese poetry. 
As introduced above, all the mentioned neural models attempt to produce 
poems with regulated rules, a consistent theme, and meaningful semantics, 
but none of them consider to represent poem theme and use it to further boost the results.
To address this issue, we propose to use a novel conditional variational autoencoder with a hybrid decoder, in which the learned latent variables combined with conditional keywords are able to convey topic information of the entire poem.

\section{Approaches}\label{model}

\begin{figure}
\centering
\includegraphics[width=0.45\textwidth]{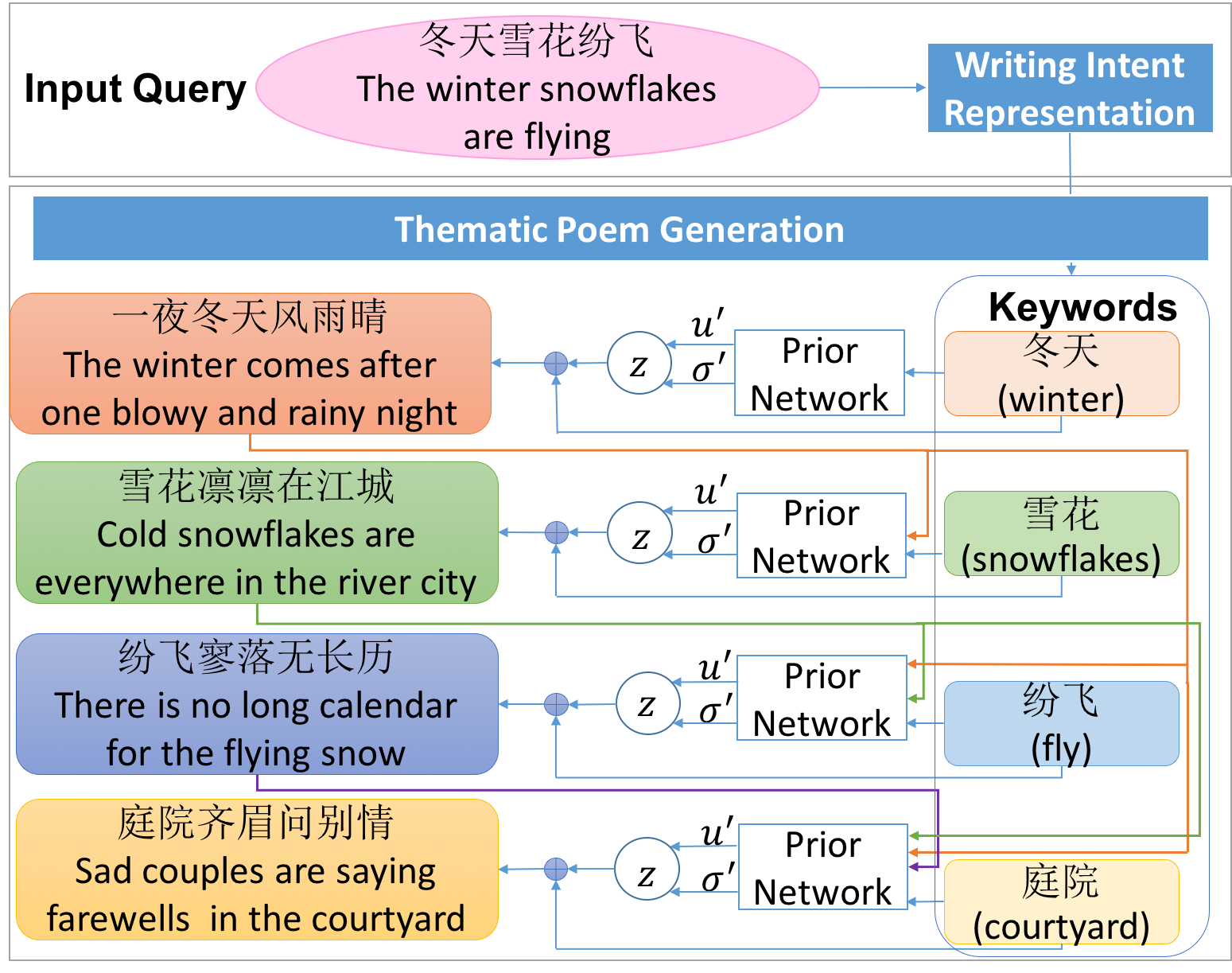}
\caption{The framework of the proposed Chinese poetry generation approach. $\oplus $ denotes the concatenation of input vectors.
}
\vspace{-3mm}
\label{figure:fig1}
\end{figure}

\subsection{Overview}
As most human poets write poems according to a sketch of ideas, we use a two-stage Chinese poem generation approach, i.e., writing intent representation and thematic poem generation. Specifically, our system can take a word, a sentence or even a document as input containing users' writing intent, and then generate rule-complied and theme-consistent poem sequentially using an improved conditional variational autoencoder. Similar work has been done in \cite{wangzhe2016chinese}, the main distinction from our work to theirs is the implemented neural model is the conditional variational autoencoders in our work.

The framework of our Chinese poetry generation approach using the proposed Conditional Variational AutoEncoder with a Hybrid Decoder (CVAE-HD) is illustrated in Fig.\ref{figure:fig1}. 
Suppose an input query \begin{CJK}{UTF8}{gbsn}``冬天雪花纷飞''\end{CJK} (``The snowflakes are flying in winter'') is given, in the writing intent representation stage, the sentence is transformed into four keywords $k_{i}$ ($i=1,2,3,4$), i.e.,  \begin{CJK}{UTF8}{gbsn}``冬天''\end{CJK} (winter),  \begin{CJK}{UTF8}{gbsn}``雪花''\end{CJK} (snowflake),  \begin{CJK}{UTF8}{gbsn}``纷飞''\end{CJK} (fly), and  \begin{CJK}{UTF8}{gbsn}``庭院''\end{CJK} (courtyard), where $k_{i}$ represents the sub-topic for the corresponding $i$th line $l_{i}$. 
In thematic poem generation stage, assuming that keywords are not enough to convey topic information for the entire poem, each line $l_{i}$ is first encoded into a latent variable $z_{i}$ to learn a distribution over potential writing intent by a prior network, and then generated by decoding from a concatenation of the learned latent variable $z_{i}$ and the extracted or expanded keyword $k_{i}$. As a result, the poem is created automatically not only by the sub-topic provided by the corresponding keyword, but also the topic messages stored in latent variables, which are learned from the current line $l_{i}$, the previously generated lines $l_{1:(i-1)}$, and the corresponding keyword $k_{i}$.
Note that the seven-character quatrain given in Fig.\ref{figure:fig1} is produced automatically from our generation system.

\subsection{Writing Intent Representation}
Due to the fact that each line of a quatrain consists of five or seven characters, we hypothesize that the sub-topic of each line can be represented by one keyword. Therefore, it is important to evaluate the importance of words extracted from the input query provided by users. We use TextRank \cite{kingma2014adam} 
to measure the importance of different words.

In the graph of TextRank, a vertex represents each candidate word and edges between two words indicate their co-occurrence, where the edge weight is set according to the total count of co-occurrence strength between these two words.
The TextRank score $S(V_{i})$ is computed iteratively until convergence according to the following equation:
\begin{equation}\label{equ0}
S(V_{i}) = (1-d) + d \sum_{V_{j} \in E(V_{i})}\frac{w_{ji}}{\Sigma _{V_{k} \in E(V_{i})}w_{jk}}S(V_{j})
\end{equation} where $w_{ji}$ is the weight of the edge between node $V_{j}$ and $V_{i}$, $E(V_{i})$ is the set of vertices connected with $V_{i}$, and $d$ is a damping factor. Empirically, the damping factor $d$ is usually set to 0.85, and the initial score of $S(V_{i})$ is set to 1.
When the number of extracted keywords from users' input query is less than the required one, we 
conduct the keyword extension in which the candidate word with the highest TextRank score is selected as the new keyword.

\subsection{Conditional Variational Autoencoders with Hybrid Decoders}\label{cvae}
For Chinese poetry generation, since most human poets create poems based on a plain outline, we believe that keywords $k_{i}$ ($i=1,2,3,4$) obtained from the first stage of our generation framework can partially represent users' writing intent, and regard them as the conditions $c$ for CVAE. 

We define the conditional distribution as $p(x, z|c) = p(x|z, c)p(z|c)$,  and set the learning target to approximate $p(z|c)$ and $p(x|z, c)$ via deep neural networks parameterized by $\theta$. 
CVAE is trained to maximize the conditional log likelihood of $x$ given $c$, meanwhile minimizing the KL regularizer between the posterior distribution $p(z|x, c)$ and a prior distribution $p(z|c)$. 
We use a recognition network $q_{\phi}(z|x,c)$  and a prior network $p_{\theta}(z|c)$ to approximate the true posterior distribution $p(z|x, c)$ and the prior distribution $p(z|c)$.
To sum up, the objective of the traditional CVAE takes the following form:
\begin{equation}\label{equ1}
\begin{split}
L(\theta ,\phi ;x,c)_{cvae} = \mathbf{E}q_{\phi}(z|x,c)[log p_{\theta}(x|z, c)] \\
-KL(q_{\phi}(z|x,c))\left |  \right | p_{\theta}(z|c))\\
\leq  log p(x|c)
\end{split}
\end{equation}

\begin{figure}
\centering
\includegraphics[width=0.45\textwidth]{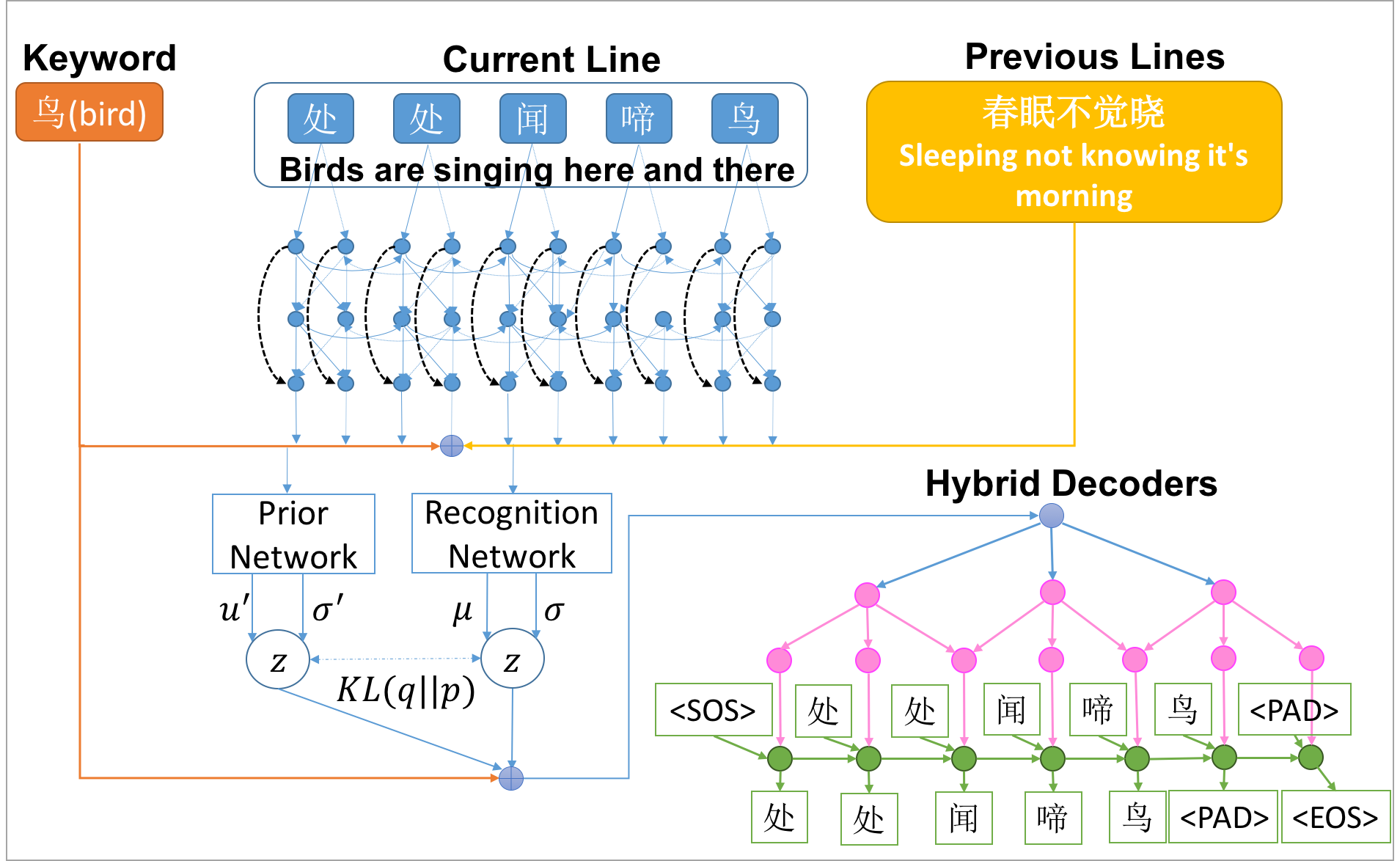}
\caption{The training procedure of the proposed Conditional Variational Autoencoders with Hybrid Decoders. The black dashed lines represent the residual connection between layers. 
}
\label{figure:fig2}
\vspace{-3mm}
\end{figure}

As shown in Eqn.\ref{equ1}, 
the generative process of $x$ can be summarized as sampling a latent variable $z$ from $p_{\theta}(z|c)$ and then generating $x$ by $p_{\theta}(x|z, c)$.
CVAE can be efficiently trained with the Stochastic Gradient Variational Bayes (SGVB) framework \cite{kingma2014adam} by maximizing the variational lower bound of the conditional log likelihood \cite{sohn2015learning}.
Fig.\ref{figure:fig2} illustrates the training procedure of our proposed Conditional Variational Autoencoders with Hybrid Decoders (CVAE-HD). 
As shown in Fig.\ref{figure:fig2}, we use a Bidirectional Recurrent Neural Network (BRNN)  \cite{kingma2014adam} with a Long Short Term Memory (LSTM) \cite{hochreiter1997long}  as an encoder to encode each concatenation of the current line $l_{i}$, the corresponding keyword $k_{i}$, and previously generated lines $l_{1:(i-1)}$ into fixed-size vectors by concatenating the last hidden states of the forward and backward RNN $h_{i} = \left [ { \overrightarrow{h_{i}}} , \overleftarrow{h_{i}} \right ]
$.  Then, $x$ can be simply represented by $h$.
We adopt multiple layers and residual connections \cite{he2016deep} between layers to learn a describable $h$ .
We suppose $z$ follows a multivariate Gaussian distribution with a diagonal covariance matrix, thus the recognition network $q_{\phi}(z|x,c) \sim \mathcal{N}(\mu , \sigma^{2} \mathbf{I} )$ and the prior network $p_{\theta}(z|c) \sim \mathcal{N}(\mu ^{'}, \sigma^{'2} \mathbf{I} )$, and then we have:

\begin{equation}\label{equ2}
\begin{bmatrix}
\mu \\ 
log(\sigma ^{2}) 
\end{bmatrix} =W_{r}\begin{bmatrix}
x\\ 
c
\end{bmatrix}+b_{r}
\end{equation}

\begin{equation}\label{equ3}
\begin{bmatrix}
\mu^{'} \\ 
log(\sigma ^{'2}) 
\end{bmatrix} = \mathbf{MLP}_{p}(c)
\end{equation} We use a reparametrization trick \cite{kingma2013auto} to sample $z$ from the recognition network $\mathcal{N}(\mu , \sigma^{2} \mathbf{I} )$ during training and $\mathcal{N}(\mu ^{'}, \sigma^{'2} \mathbf{I} )$ predicted by the prior network during testing. The initial state $s_{0} = W_{d}[z,c]+b_{d}$ is used for a RNN decoder.

Since it is easy for CVAE to ignore the latent variable $z$ when directly using an RNN decoder, inspired by \cite
{semeniuta2017hybrid}, we propose to use a hybrid decoder in CVAE as shown in Fig.\ref{figure:fig2} and name the novel CVAE as Conditional Variational Autoencoders with Hybrid Decoders (CVAE-HD). The hybrid decoder is composed of deConvolutional Neural Networks (deCNN) \cite{radford2015unsupervised,gulrajani2016pixelvae} and recurrent neural networks. The reason we introduce deCNN as a part of the decoder in CVAE is to build the connection of each element in $x$ with the learned latent variable $z$. Then, the probability of the generated sequence $x$ can be represented as $P(x_1,...,x_{n}|z, c) = \prod _{i} P(x_i|z, c)$. However, it is hard for a fully feed-forward architecture to learn the sequential information between the element in $x$. A multi-layer LSTM decoder similar with the encoder is added on top of deCNN layers to model $P (x_1 , . . . , x_n |z,c) = \prod _{i} P (x_i|x_{i-1}, . . . , x_1,z,c)$.

\subsection{Optimization}\label{sec:opt}
Although CVAE has achieved impressive results in image generation, it is non-trivial to adapt it to natural language generators due to the \textit{vanishing latent variable problem}. KL annealing \cite{bowman2015generating} gradually increasing the weight of the KL term from 0 to 1 during training plays a powerful role in dealing with this problem. 
Another solution word drop decoding, which sets a certain percentage of the target words to 0, may hurt the performance when the drop rate is too high. Thus, we adopt KL annealing instead of word drop decoding during training for CVAE.

Beyond that, we propose an auxiliary solution to help further solve the above problem, i.e., we add an additional deCNN reconstruction loss term to Eqn.\ref{equ1} and regularize it with a weighting parameter $\alpha$. Therefore, the loss function of our proposed CVAE-HD can be represented as below:
\begin{equation}\label{equ4}
\begin{split}
L_{cvae-hd} = L_{cvae} + \alpha L_{dcnn}
\end{split}
\end{equation} in which the second term is computed from the activations of the last deconvolutional layer $L(\theta ,\phi ;x,c)_{dcnn} = \mathbf{E}q_{\phi}(z|x,c)[log p_{\theta}(x|z, c)]$.

Since we use the combination of both keywords extracted from users' query and the latent variable $z$ learned from CVAE to represent poetry theme, the representation of keywords is the key for the performance to some extent. Therefore, we try to mine the nature of quatrains to obtain a good representation of poetry word. We notice that for some lines in quatrains, mostly the third and the fourth line, corresponding characters from the same position in these two lines often match each other by certain constraints on semantic and/or syntactic relatedness. 

Taking two lines \begin{CJK}{UTF8}{gbsn}``千山鸟飞绝 (A thousand mountains without birds flying)，万径人踪灭 (Ten thousand paths without a footprint)''\end{CJK} of the famous five-quatrain \begin{CJK}{UTF8}{gbsn}``江雪''\end{CJK} (River Snow) as an example, the characters \begin{CJK}{UTF8}{gbsn}``千''\end{CJK} (thousand) and \begin{CJK}{UTF8}{gbsn}``万''\end{CJK} (ten thousand) both represent numbers, meanwhile  \begin{CJK}{UTF8}{gbsn}``绝''\end{CJK} (gone) and \begin{CJK}{UTF8}{gbsn}``灭''\end{CJK} (disappeared) both deliver similar meanings of nonexistence.  
Even though the constraints of quatrains are not as strict as the Chinese antithetical couplets \cite{yan2016chinese}, we propose to initialize the word-embedding vectors using an Augmented Word2Vec model (AW2V) to further enhance the rhythm and symmetry delivered in poems.
This model adds vertical slices of poems as additional sentences
to the training data based on the word2vec \cite{mikolov2013efficient}.
AW2V is not only able to boost the optimization of CVAE and improve the rule-consistency of generated poems, but also can be used to search characters representing similar semantic meanings in Chinese poems. 

\begin{table*}
\centering
\begin{tabular}{|c|c|c|c|c|c|c|c|} \hline
Approach  & PPL & NNL(KL) & BLEU-1 & BLEU-2 & BLEU-3 & BLEU-4 & RES \\ \hline\hline
Ground truth   & - & -& - & - & - & - & 0.8975 \\ \hline
AS2S   & 55.5255 & 4.0168(-) &  0.4296 & 0.3596 & 0.3045 & 0.2640 & 0.5046\\ \hline
AS2S+AW2V   & 54.1752 & 3.9922(-)& 0.4319 & 0.3625 & 0.3076 & 0.2669 & 0.5076\\ \hline
CVAE & 52.1154 & 3.9535(0.0083) & 0.4366 & 0.3605 & 0.3046 & 0.2637  & 0.5137\\ \hline
CVAE+AW2V & 52.0516 & 3.9522(0.0126) & 0.4382 &  0.3630 & 0.3073  &  0.2663  & 0.5149 \\ \hline
CVAE-HD+AW2V & \textbf{51.7236} & \textbf{3.9459(0.0163)} & \textbf{0.4405} & \textbf{0.3644} & \textbf{0.3084} & \textbf{0.2672} & \textbf{0.5245}\\ \hline
\end{tabular}
\caption{The language modeling results and the performance on automatic metrics on the test dataset. We report negative log likelihood (NLL) and perplexity (PPL), and the KL component of NLL is given in parentheses. Note that the reported BLEU scores are normalized  
to [0, 1]. The mean score of the proposed Rhythm Score Evaluation (RSE) is reported in the last column. 
The corresponding best scores are shown in bold. 
}
\label{table:all-measure}
\vspace{-3mm}
\end{table*}

\section{Experimental Setup}\label{exp}

\subsection{Dataset}\label{sec:data}
Two large-scale datasets are used in our experiments. The first dataset is a Chinese poem corpus (CPC) containing 284,899 traditional Chinese poems in various genres, including Tang quatrains, Song Iambics, Yuan Songs, and Ming and Qing poems. We use this dataset to train the word-embedding for Chinese characters.
Since we focus on generating quatrains which have four lines with the same length of five or seven characters in each line, we filter 76,305 quatrains from CPC, named as Chinese quatrain corpus (CQC), to train the neural network model. Specifically, we randomly choose 2,000 poems for validation, 1,000 poems for testing, and other non-overlap ones for training. We segment all the poems into words and calculate the TextRank score for each word. Then, the word with the highest TextRank score is selected as the keyword for each line so that each quatrain owns four keywords. 

\subsection{Training}\label{sec:training}
We choose the 6,000 most frequently used characters as the vocabulary.  
The dimension of word-embedding vectors is set to 128.
The recurrent hidden layers of the encoder and the RNN part of the hybrid decoder contain 128 hidden units, and the number of layers is both set to 4.
We use 3 deconvolutional layers with the ReLU non-linearity in the deCNN part of the hybrid decoder. The kernel size is set to 3 and the stride is 2. The number of feature maps is [256, 128, 64] for each layer respectively. 
The weighting parameter $\alpha$ is set to 0.6.
We use 64-dimensional latent variables.
Parameters of our model were randomly initialized over a uniform distribution with support [-0.08,0.08].
The model is trained using the AdaDelta algorithm \cite{zeiler2012adadelta}, where the mini-batch is set to 64 and the learning rate is 0.001. The dropout technique \cite{srivastava2014dropout} is also adopted and the dropout rate is set to 0.2.
The perplexity value on the validation set is used for the early stop of training to avoid an overfitting learned model. 

\section{Evaluation}
Generally, it is difficult to judge the quality of poems generated by computers. 
We conduct both automatic and human evaluation to verify the feasibility and availability of our proposed Chinese poetry generation approach. 

For the comparative approach, we mainly compare our proposed approach with the attention-based sequence-to-sequence model (AS2S) presented in \cite{wangzhe2016chinese}, which has been proved to be capable of generating different genres of Chinese poems. The reasons we choose AS2S to compare rather than others can be summarized into two aspects. First, this model has been fully compared with other previous methods such as SMT, RNNLM, RNNPG, and ANMT, and proved better than all of them. Second, the first generation phase of our proposed approach, i.e., writing intent representation, is similar to the procedure introduced in \cite{wangzhe2016chinese} while the second phase is completely different. Therefore, through comparing our framework with theirs, we can inspect the effects of our proposed Conditional Variational AutoEncoders with Hybrid Decoders (CVAE-HD). 

\subsection{Automatic Evaluation}

\subsubsection{Poetry Modeling Results}\label{sec:ppl}
The language modeling results on the test dataset of CQC are shown in Table \ref{table:all-measure}, in which the reconstruction perplexity (PPL), negative log likelihood (NLL) and the KL component (KL) are reported. In addition to this, BLEU evaluation method \cite{papineni2002bleu}, which is famous for the evaluation the quality of the text, is also reported in Table \ref{table:all-measure}. We use BLEU-1 to 4, and normalize them to $[0,1]$ scale.

Since Chinese quatrains have strict regulations and should follow particular tonal and structural rules, we propose a new Rhythm Score Evaluation (RSE) to automatically measure the rule-consistency of poems. We define RSE as 
\begin{equation}\label{equ_rhythm}
Rhy(l) = \left\{\begin{matrix}
0, & cnt(l)\notin   \left \{ 5, 7 \right \}\\ 
0.5, & rule(l)\in T~or~R\\ 
1, & rule(l)\in T~and~R
\end{matrix}\right.
\end{equation} where $l$ represents each line of poems, $cnt(l)$ is the number of characters of $l$, $rule(l)$ is the rule of $l$, and $T$ and $R$ represent the set of tonal patterns and rhyming patterns severally. 
The results of mean score in terms of RSE are demonstrated in the last column of Table \ref{table:all-measure}. 
A higher mean score indicates approaches owning better capability of generating poems with regulated rhythm and structure. 
Note that the ground truth represents the humanly written poems in the test dataset. 

From Table \ref{table:all-measure}, we can find that, compared with AS2S as the baseline, both CVAE+AW2V and CVAE-HD+AW2V outperform in terms of all the metrics. Note that we represent the approach using our proposed augmented word2vec model (AW2V) by appending a plus sign to the original method, e.g., AS2S+AW2V. Compared with AS2S+AW2V and AS2S, and CVAE+AW2V and CVAE, the improvement by adding AW2V can be found in both AS2S and CVAE, which demonstrates the advantage of AW2V in the optimization for poetry modeling. 
Beyond this, CVAE-HD, the proposed novel CVAE with a hybrid decoder, outperforms CVAE especially in the terms of KL. This proves that the hybrid decoder can relieve the pressure of vanishing latent variables to a certain extent.
We notice that due to the simplification of poetry, although KL annealing and our proposed hybrid decoder are all adopted during training, it is harder for poetry generation to tackle the vanishing latent variable problem than general natural language generation. 

\begin{CJK}{UTF8}{gbsn}
\begin{table*}
\centering
\begin{tabular}{|c|c|} \hline
蜡烛今宵尽，& 寒梅寂寞是无家，\\
Burned candle flickered at dawn,& Lonely, a plum blossom is homeless,\\
残灯隔户人。& 未折惊心待岁华。\\
A dim light shone on the man home alone. & Apprehensive, the ephemeral beauty will wither. \\
衣襟因酒别，& 折得一枝凝瘦骨，\\
Drinking, my body and mind flowed, & Hey, don’t worry. I will pick a twig in the woods,\\
何况雪中人。& 砌间长笛有梅花。\\
While you were fainting in the snow. & And house it in my flute. \\
\hline
\end{tabular}
\caption{The five/seven-character generated quatrain based on the given query ``蜡烛 (candle)''/``梅 (wintersweet)''.}
\label{poem:twoexamples}
\end{table*}
\end{CJK}

\subsubsection{Poetry Character Similarity}\label{sec:w2v}
We measure the similarity of poetry characters to verify the superiority of our proposed AW2V model over the original word2vec (W2V) one.

Taking the poetry \begin{CJK}{UTF8}{gbsn}``江雪''\end{CJK} (River Snow) mentioned in Section \ref{sec:opt} as an instance, the similarity between \begin{CJK}{UTF8}{gbsn}千 (thousand)\end{CJK} and \begin{CJK}{UTF8}{gbsn}万 (ten thousand)\end{CJK} using AW2V is 0.4389, while 0.4039 using W2V. It is worth noticing that \begin{CJK}{UTF8}{gbsn}绝 (gone)\end{CJK} and \begin{CJK}{UTF8}{gbsn}灭 (disappeared)\end{CJK} get a 0.2745 similarity score in AW2V model, while only get 0.0205 in W2V. 
Beyond that, we can use AW2V to search similar words. For instance, if we search similar words for \begin{CJK}{UTF8}{gbsn}年 (year)\end{CJK}, we obtain \begin{CJK}{UTF8}{gbsn}旬 (ten days)\end{CJK}, \begin{CJK}{UTF8}{gbsn}番 (multiple times)\end{CJK}, and \begin{CJK}{UTF8}{gbsn}时 (time)\end{CJK} which are all time-related Chinese characters. 

\subsection{Human Evaluation}\label{sec:human}
\subsubsection{Online Test}\label{sec:onlinetest}

\begin{figure}
\centering
\includegraphics[width=0.3\textwidth]{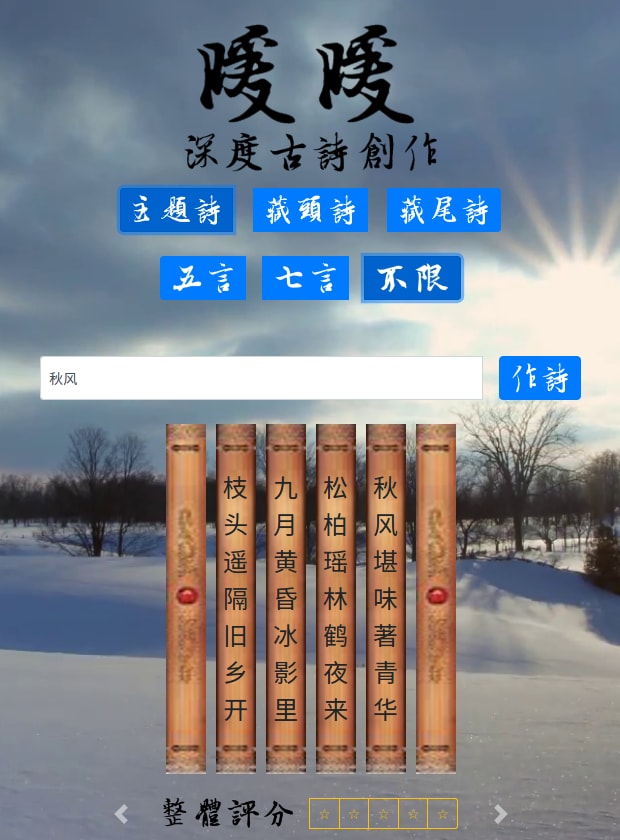}
\caption{The website interface of our poetry generation system in which users are asked to input the query and rate the generated poem based on certain metrics [best viewed in color].}
\label{figure:fig4}
\vspace{-3mm}
\end{figure}

We build a web-based environment for our poetry generation system whose interface is illustrated in Fig.\ref{figure:fig4}.
Using this website, users can input any arbitrary query as the topic to generate a computer-written poem using our proposed CVAE-HD.

We invited users who are well-educated and have a great passion for poetry writing to participate our human evaluation. 
All the participants are asked to rate poems in the score range $[1,5]$ based on five subjective evaluation metrics
including \textit{Readability} (if the sentences read smoothly and fluently), \textit{Consistency} (if the entire poem delivers a consistent theme), \textit{Aesthetic} (if the quatrain stimulates any aesthetic feeling), \textit{Evocative} (if the quatrain expresses meaningful emotion), and \textit{Overall} (if the quatrain is overall well written).

As shown in Fig.\ref{figure:fig4}, the illustrated example contains the user input query \begin{CJK}{UTF8}{gbsn}``秋风''\end{CJK} (Autumn Wind) and the corresponding generated poem \begin{CJK}{UTF8}{gbsn}``秋风堪味著青华 (The lush growth of trees colors the early autumn wind)，松柏瑶林鹤夜来 (A group of cranes flies through the forest in the evening)。九月黄昏冰影里 (It is only September and the frost is shivering at dusk)， 枝头遥隔旧香来 (The branches remind me of the distance separating us)。''\end{CJK}. 
We notice that the generated poem not only contains the related words of ``autumn wind,''  e.g., ``cranes'' and ``September,'' but also delivers consistently gloomy theme and emotion.  This poem, moreover, intuitively conforms to the tonal and structural rules of quatrains.

Up to now, we collect 139 quatrains based on all the random queries input by the human-evaluation participants. On average, we obtain a 3.43 \textit{Readability} score, a 3.15  \textit{Consistency} score, a 3.26 \textit{Aesthetic} score, a 3.16 \textit{Evocative} score, and a 3.22 \textit{Overall} score. Among all the generated poems, 73.42\% of them receive an \textit{Overall} score no less than 3.

We also give some other specific examples based on various given queries.
Two quatrains produced by our poetry generation system are shown in Table \ref{poem:twoexamples}.



\section{Conclusions}\label{conc}

In this work, we have studied poetry generation. 
We present a two-step generation approach including writing intent representation and thematic poem generation to imitate the poem creation process by human poets.
We have proposed a conditional variational autoencoder with a hybrid decoder to mine the implicit topic information contained within poems lines. An augmented word2vec model has also been proposed to further enhance the rhythm and symmetry delivered in poems and improve the training procedure.
The generative neural model can incorporate more flexibility to represent the theme message by learning latent variables.

We conduct the experiments on several evaluation metrics and compare our proposed approach with some existing ones. 
Experimental results demonstrate that our proposed poetry generation approach can produce satisfying quatrains with regulated rules and consistent themes. 
Our proposed conditional variational autoencoder with a hybrid decoder has been proved to outperform the attention-based sequence-to-sequence model.
 
Currently, we are working on using reinforcement learning to further improve the poetry quality
 and generating different literature, such as lyrics and compositions.

\section*{Acknowledgments}
This work is partially supported by China's National Key Research and Development Program under grant 2016YFB1000902 and 2018YFB1003202, Beijing Advanced Innovation Center for Imaging Technology BAICIT-2016031, Ningbo Science and Technology Innovation team No. 2014B82014, Canada's NSERC OGP0046506, and Canada Research Chair Program.

\bibliographystyle{named}
\bibliography{poem-ref}

\end{document}